\pdfoutput=1

\documentclass[11pt]{article}

\usepackage{acl}
\usepackage{times}
\usepackage{latexsym}

\usepackage[T1]{fontenc}

\usepackage[utf8]{inputenc}

\usepackage{microtype}
\usepackage{multirow}
\usepackage{multicol}
\usepackage{makecell}
\usepackage[skins]{tcolorbox}
\usepackage{amsmath}
\usepackage{amsfonts}
\usepackage{adjustbox}
\usepackage{bm}
\usepackage{enumitem}
\usepackage{circledsteps}

\def \margin {4pt}
\def \fmargin {-10pt}

\title{\textsc{ORGan}: Observation-Guided Radiology Report Generation via Tree Reasoning}
\author{Wenjun Hou$^{1,2}$, Kaishuai Xu$^{1\ast}$, Yi Cheng$^{1\ast}$, Wenjie Li$^{1\dagger}$, Jiang Liu$^{2\dagger}$ \\
$^1$Department of Computing, The Hong Kong Polytechnic University, HKSAR, China \\
$^2$Research Institute of Trustworthy Autonomous Systems and \\Department of Computer Science and Engineering, \\
Southern University of Science and Technology, Shenzhen, China \\
\texttt{houwenjun060@gmail.com} \\
\texttt{\{kaishuaii.xu, alyssa.cheng\}@connect.polyu.hk} \\
\texttt{cswjli@comp.polyu.edu.hk, liuj@sustech.edu.cn}
}

\begin{document}
\maketitle
\begingroup\def\thefootnote{$\ast$}\footnotetext{Equal Contribution.}\endgroup
\begingroup\def\thefootnote{$\dagger$}\footnotetext{Corresponding authors.}\endgroup
\begin{abstract} 
This paper explores the task of radiology report generation, which aims at generating free-text descriptions for a set of radiographs. One significant challenge of this task is how to correctly maintain the consistency between the images and the lengthy report. Previous research explored solving this issue through planning-based methods, which generate reports only based on high-level plans. However, these plans usually only contain the major observations from the radiographs (e.g., lung opacity), lacking much necessary information, such as the observation characteristics and preliminary clinical diagnoses. To address this problem, the system should also take the image information into account together with the textual plan and perform stronger reasoning during the generation process. In this paper, we propose an \underline{O}bservation-guided radiology \underline{R}eport \underline{G}ener\underline{\textsc{a}}tio\underline{\textsc{n}} framework (\textsc{ORGan}). It first produces an observation plan and then feeds both the plan and radiographs for report generation, where an observation graph and a tree reasoning mechanism are adopted to precisely enrich the plan information by capturing the multi-formats of each observation. Experimental results demonstrate that our framework outperforms previous state-of-the-art methods regarding text quality and clinical efficacy.\footnote{Our code is available at \url{https://github.com/wjhou/ORGan}.} 
\end{abstract}
\section{Introduction}
Radiology reports, which contain the textual description for a set of radiographs, are critical in the process of medical diagnosis and treatment. Nevertheless, the interpretation of radiographs is very time-consuming, even for experienced radiologists (5-10 minutes per image). Due to its large potential to alleviate the strain on the healthcare workforce, automated radiology report generation \cite{topdown,att2in} has attracted increasing research attention.
\begin{figure}[t]
	\centering
    \setlength\belowcaptionskip{\fmargin}
    \includegraphics[width=1.0\linewidth]{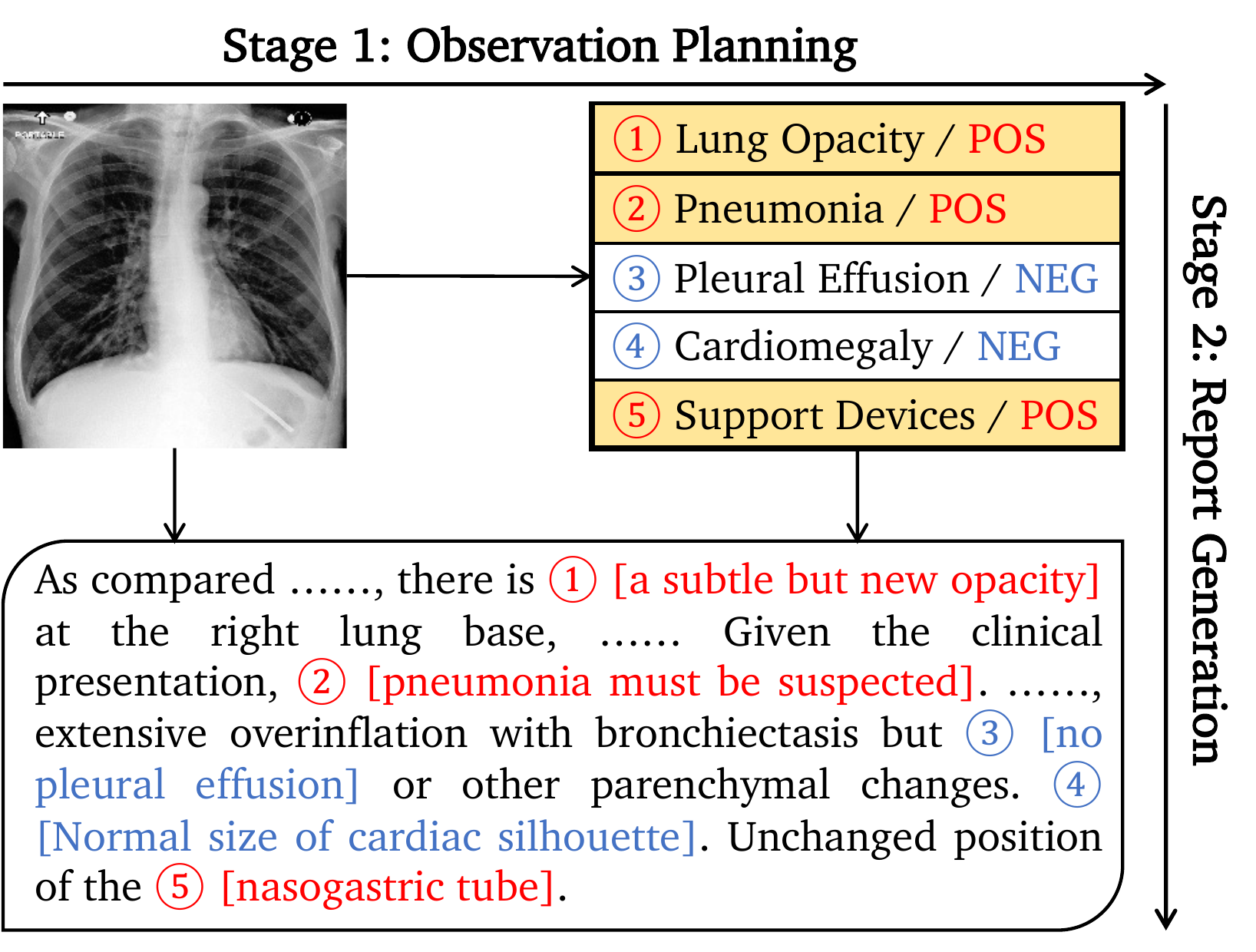}
	\caption{Our proposed framework contains two stages, including the observation planning stage and the report generation stage. {\color{red} Red} color denotes positive observations, while {\color{blue} Blue} color denotes negative observations.}
    \label{figure: example}
\end{figure}

One significant challenge of this task is how to correctly maintain the consistency between the image and the lengthy textual report. Many previous works proposed to solve this through planning-based generation by first concluding the major observations identified in the radiographs before the word-level realization \cite{coatt,aligntransformer,m2tr,coplan}. Despite their progress, these methods still struggle to maintain the cross-modal consistency between radiographs and reports. A significant problem within these methods is that, in the stage of word-level generation, the semantic information of observations and radiographs is not fully utilized. They either generate the report only based on the high-level textual plan (i.e., major observations) or ignore the status of an observation (i.e., positive, negative, and uncertain), which is far from adequate. The observations contained in the high-level plan are extremely concise (e.g., lung opacity), while the final report needs to include more detailed information, such as the characteristics of the observation (e.g., a subtle but new lung opacity) and requires preliminary diagnosis inference based on the observation (e.g., lung infection must be suspected). In order to identify those detailed descriptions and clinical inferences about the observations, we need to further consider the image information together with the textual plan, and stronger reasoning must be adopted during the generation process.

In this paper, we propose \textsc{ORGan}, an \underline{O}bservation-guided radiology \underline{R}eport \underline{G}ener\underline{\textsc{a}}tio\underline{\textsc{n}} framework. Our framework mainly involves two stages, i.e., the observation planning and the report generation stages, as depicted in Figure \ref{figure: example}. In the first stage, our framework produces the observation plan based on the given images, which includes the major findings from the radiographs and their statuses (i.e., positive, negative, and uncertain). In the second stage, we feed both images and the observation plan into a Transformer model to generate the report. Here, a tree reasoning mechanism is devised to enrich the concise observation plan precisely. Specifically, we construct a three-level observation graph, with the high-level observations as the first  level, the observation-aware n-grams as the second level, and the specific tokens as the third level. These observation-aware n-grams capture different common descriptions of the observations and serve as the component of observation mentions. Then, we use the tree reasoning mechanism to capture observation-aware information by dynamically aggregating nodes in the graph. 

In conclusion, our main contributions can be summarized as follows: 
\begin{itemize}
    \item We propose an \underline{O}bservation-guided radiology \underline{R}eport \underline{G}ener\underline{\textsc{a}}tio\underline{\textsc{n}} framework (\textsc{ORGan}) that can maintain the clinical consistency between radiographs and generated free-text reports.
    \item To achieve better observation realization, we construct a three-level observation graph containing observations, n-grams, and tokens based on the training corpus. Then, we perform tree reasoning over the graph to dynamically select observation-relevant information.
    \item We conduct extensive experiments on two publicly available benchmarks, and experimental results demonstrate the effectiveness of our model. We also conduct a detailed case analysis to illustrate the benefits of incorporating observation-related information.
\end{itemize}
\section{Methodology}
\subsection{Overview of the Proposed Framework}
Given an image $X$, the probability of a report $Y=\{y_1,\dots,y_T\}$ is denoted as $p(Y|X)$. Our framework decomposes $p(Y|X)$ into two stages, where the first stage is observation planning, and the second stage is report generation. Specifically, observations of an image $Z=\{z_1,\dots,z_L\}$ are firstly produced, modeled as $p(Z|X)$. Then, the report is generated based on the observation plan and the image, modeled as $p(Y|X, Z)$. Finally, our framework maximizes the following probability:
\begin{equation*}
\setlength{\belowdisplayskip}{\margin}
\setlength{\abovedisplayskip}{\margin}
    p(Y|X)\propto\underbrace{p(Z|X)}_{\text{Stage 1}}\underbrace{p(Y|X,Z)}_{\text{Stage 2}}.
\end{equation*}

\begin{figure*}[t]
	\centering
    \setlength\belowcaptionskip{\fmargin}
    \includegraphics[width=0.88\linewidth]{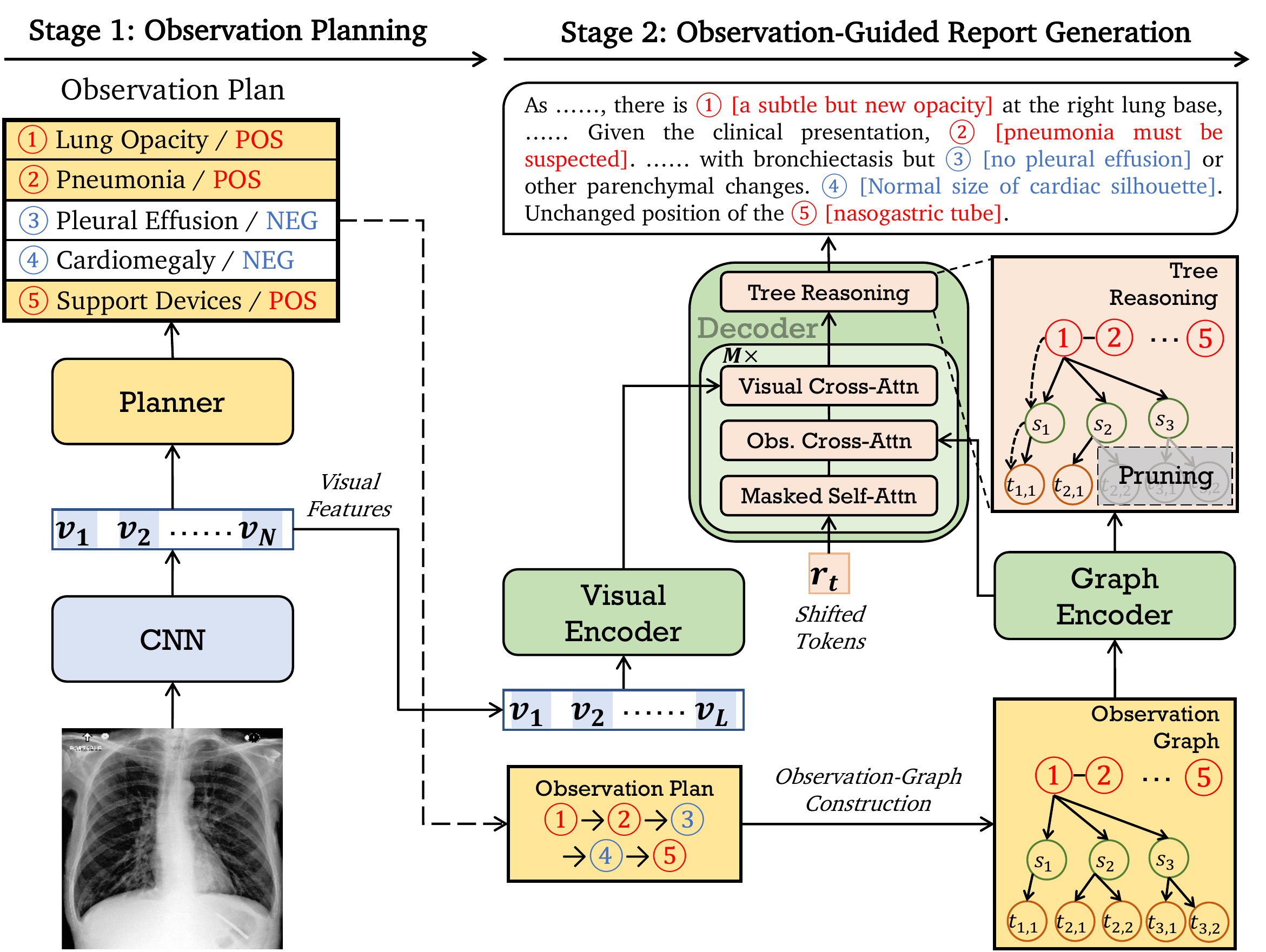}
	\caption{The overall framework of \textsc{ORGan} (``\emph{Obs. Cross-Attn}'' in the decoder refers to the observation-related cross-attention module). }
    \label{figure: overall_framework}
\end{figure*}
\subsection{Observation Plan Extraction and Graph Construction} 
\textbf{Observation Plan Extraction.} There are two available tools for extracting observation labels from reports, which are CheXpert \cite{chexpert} and CheXbert \cite{chexbert}. We use CheXbert\footnote{\url{https://github.com/stanfordmlgroup/CheXbert}} instead of CheXpert because the former achieved better performance. To extract the observation plan of a given report, we first adopt the CheXbert to obtain the observation labels within 14 categories $C=\{C_1,\dots, C_{14}\}$ as indicated in \citet{chexpert}. More details about the distribution of observation can be found in Appendix \ref{appendix:observation}. The label (or status) of each category belongs to \textit{Present}, \textit{Absent}, and \textit{Uncertain}, except the \textit{No Finding} category, which only belongs to \textit{Present} and \textit{Absent}. To simplify the observation plan and emphasize the abnormalities presented in a report, we regard \textit{Present} and \textit{Uncertain} as Positive and \textit{Absent} as Negative. Then, observations are divided into a positive collection $C\text{/POS}$ and a negative collection $C\text{/NEG}$, and each category with its corresponding label is then converted to its unique observation $C_i\text{/POS} \in C\text{/POS}$ or $C_i\text{/NEG} \in C\text{/NEG}$, resulting in 28 observations. For example, as indicated in Figure 1, the report presents \textit{Lung Opacity} while \textit{Cardiomegaly} is absent in it. These categories are converted to two observations: \textit{Lung Opacity}/POS and \textit{Cardiomegaly}/NEG. Then, we locate each observation by matching mentions in the report and order them according to their positions. These mentions are either provided by \citet{chexpert}\footnote{\url{https://github.com/stanfordmlgroup/chexpert-labeler}} or extracted from the training corpus (i.e., n-grams), as will be illustrated in the following part. Finally, we can obtain the image's observation plan $Z=\{z_1,\dots,z_L\}$.

\noindent\textbf{Tree-Structured Observation Graph Construction}. Since observations are high-level concepts that are implicitly related to tokens in reports, it could be difficult for a model to realize these concepts in detailed reports without more comprehensive modeling. Thus, we propose to construct an observation graph by extracting observation-related n-grams as the connections between observations and tokens for better observation realization. Specifically, it involves two steps to construct such a graph: (1) n-grams extraction, where $n \in \left[1,4 \right]$ and (2) <observation, n-gram> association. Following previous research \cite{ngram_pmi,style_pmi}, we adopt the pointwise mutual information (PMI) \cite{pmi} to fulfill these two steps, where a higher PMI score implies two units with higher co-occurrence:
\begin{equation*}
\setlength{\belowdisplayskip}{\margin}
\setlength{\abovedisplayskip}{\margin}
    \text{PMI}(\bar{x},\hat{x})=log\frac{p(\bar{x},\hat{x})}{p(\bar{x})p(\hat{x})}.
\end{equation*}
For the first step, we extract n-gram units $S=\{s_1,\dots,s_{|s|}\}$ based on the training reports. Given two adjacent units $\bar{x}$ and $\hat{x}$ of a text sequence, a high PMI score indicates that they are good collection pairs to form a candidate n-gram $s_*$, while a low PMI score indicates that these two units should be separated. For the second step, given a predefined observation set $O=\{z_1,\dots,z_{|O|}\}$, we extract the observation-related n-gram units with $\text{PMI}(z_i,s_j)$, where $z_i$ is the $i$-th observation, $s_j$ is the $j$-th n-gram, and $p(z_i, s_j)$ is the frequency that an n-gram $s_j$ appears in a report with observation $z_i$ in the training set. Then, we can obtain a set of observation-related n-grams $s^{z}=\{s^z_1,\dots,s^z_k\}$, where $s^z_j=\{t^z_{j,1},\dots,t^z_{j,n}\}$, and tokens in n-grams form the token collection $T=\{t_1,\dots,t_{|T|}\}$. Note that we remove all the stopwords in $T$, using the vocabulary provided by NLTK\footnote{\url{https://www.nltk.org/}}. Finally, for each observation, we extract the top-K n-grams as the candidates to construct the graph, which contains three types of nodes $V=\{Z, S, T\}$. We list part of the n-grams in Appendix \ref{appendix:ngram}. After extracting relevant information from the training reports, we construct an observation graph $G=<V, E>$ by introducing three types of edges $E=\{E_1, E_2, E_3\}$: 
\begin{itemize}[noitemsep,topsep=2pt]
    \item $E_1$: This undirected edge connects two adjacent observations in an observation plan (i.e., <$z_i$, $z_{i+1}$>).
    \item $E_2$: This directed edge connects an observation and an n-gram (i.e., <$z_i$, $s_j$>).
    \item $E_3$: This directed edge connects an n-gram with its tokens (i.e., <$s_j$, $t_k$>).
\end{itemize}

\subsection{Visual Features Extraction}
Given an image $X$, a CNN and an MLP layer are first adopted to extract visual features $\bm X$:
\begin{equation*}
\setlength{\belowdisplayskip}{\margin}
\setlength{\abovedisplayskip}{\margin}
    \begin{split}
        \bm X&=\{\bm x_1,\dots,\bm x_N\}=\text{MLP}(\text{CNN}(X)), \\
    \end{split}
\end{equation*}
where $\bm x_i \in \mathbb{R}^h$ is the $i$-th visual feature.

\subsection{Stage 1: Observation Planning}
The output of observation planning is an observation sequence, which is the high-level summarization of the radiology report, as shown on the left side of Figure \ref{figure: overall_framework}. While examining a radiograph, a radiologist must report positive observations. However, only part of the negative observations will be reported by the radiologist,
depending on the overall conditions of the radiograph (e.g., co-occurrence of observations or the limited length of a report). Thus, it is difficult to plan without considering the observation dependencies (i.e., label dependencies). Here, we regard the planning problem as a generation task and use a Transformer encoder-decoder for observation planning:
\begin{equation*}
\setlength{\belowdisplayskip}{\margin}
\setlength{\abovedisplayskip}{\margin}
    \begin{split}
        \bm h^v=\{\bm h^v_1,\dots,\bm h^v_N\}&=\text{Encoder}_p(\bm X), \\
        \bm z_l&=\text{Decoder}_p(\bm{h}^v,\bm z_{<l}), \\
        p(z_l|X,Z_{< l})&=\text{Softmax}(\bm{W}_z\bm{z}_l+\bm{b}_z), \\
    \end{split}
\end{equation*}
where $\bm h^v_i \in \mathbb{R}^h$ is the $i$-th visual hidden representation, $\text{Encoder}_p$ is the visual encoder, $\text{Decoder}_p$ is the observation decoder, $\bm z_* \in \mathbb{R}^h$ is the decoder hidden representation, $\bm W_z \in \mathbb{R}^{|O|\times h}$ is the weight matrix, and $\bm{b}_z \in \mathbb{R}^{|O|}$ is the bias vector. Then the planning loss $\mathcal{L}_p$ is formulated as:
\begin{equation*}
\setlength{\belowdisplayskip}{\margin}
\setlength{\abovedisplayskip}{\margin}
    \begin{split}
        \mathcal{L}_p&=-\sum^L_{l=1} w_l \log p(z_l|X,Z_{< l})\\
        w_l&=\left\{
        \begin{aligned}
            &1+\alpha &\text{if}\: z_l \in C\text{/POS}, \\
            &1 &\text{otherwise}. \\
        \end{aligned}
        \right. \\
    \end{split}
\end{equation*}
By increasing $\alpha$, the planner gives more attention to abnormalities.
Note that the plugged $\alpha$ is applied to positive observations and \textit{No Finding}/NEG instead of \textit{No Finding}/POS.

\subsection{Stage 2: Observation-Guided Report Generation}
\textbf{Observation Graph Encoding}. We use a Transformer encoder to encode the observation graph constructed according to Section 2.2. To be specific, given the observation graph $G$ with nodes $V=\{Z,S,T\}$ and edges $E=\{E_1,E_2,E_3\}$, we first construct the adjacency matrix $\hat{A}=A+I$ based on $E$. Then, $V$ and $\hat{A}$ are fed into the Transformer for encoding. Now $\hat{A}$ serves as the self-attention mask in the Transformer, which only allows nodes in the graph to attend to connected neighbors and itself. To incorporate the node type information, we add a type embedding $\bm{P} \in \mathbb{R}^h$ for each node representation:
\begin{equation*}
\setlength{\belowdisplayskip}{\margin}
\setlength{\abovedisplayskip}{\margin}
    \begin{split}
        \bm{N}&=\text{Embed}(V)+\bm{P}, \\
        \bm V=\{\bm Z,\bm S,\bm T\}&=\text{Encoder}_g(\bm{N},\hat{A}), \\
    \end{split}
\end{equation*}
where Embed$(\cdot)$ is the embedding function, and $\bm{N}\in \mathbb{R}^h$ represents node embeddings. For observation nodes, $\bm{P}$ denotes positional embeddings, and for n-gram and token nodes, $\bm{P}$ represents type embeddings. $\bm{Z}$, $\bm{S}$, and $\bm{T} \in \mathbb{R}^h$ are encoded representations of observations, n-grams, and tokens, respectively.

\noindent\textbf{Vision-Graph Alignment}. As an observation graph may contain irrelevant information, it is necessary to align the graph with the visual features. Specifically, we jointly encode visual features $\bm{X}$ and token-level node representations $\bm{T}$ so that the node representations can fully interact with the visual features, and we prevent the visual features from attending the node representations by introducing a self-attention mask M:
\begin{equation*}
\setlength{\belowdisplayskip}{\margin}
\setlength{\abovedisplayskip}{\margin}
    \begin{split}
        [\bm{h}^v,\bm{T}^A]&=\text{Encoder}_u([\bm{X},\bm{T}],\text{M}), \\
    \end{split}
\end{equation*}
where $\bm{h}^v, \bm{T}^A \in \mathbb{R}^h$ are the visual representation and the aligned token-level node representations, respectively.

\noindent\textbf{Observation Graph Pruning}. After aligning visual features and the observation graph, we prune the graph by filtering out irrelevant nodes. The probability of keeping a node is denoted as:
\begin{equation*}
\setlength{\belowdisplayskip}{\margin}
\setlength{\abovedisplayskip}{\margin}
    \begin{split}
        p(1|\bm{T}^A)&=\text{Sigmoid}(\bm{W}_d\bm{T}^A+ b_d), \\
    \end{split}
\end{equation*}
where $\bm{W}_d \in \mathbb{R}^{1 \times h}$ is the learnable weight and $b_d \in \mathbb{R}$ is the bias. We can optimize the pruning process with the following loss:
\begin{equation*}
\setlength{\belowdisplayskip}{\margin}
\setlength{\abovedisplayskip}{\margin}
    \begin{split}
        \mathcal{L}_d&=[-\beta\cdot d\log p(1|\bm{T}^A)\\
        &-(1-d)\log (1-p(1|\bm{T}^A))], \\
    \end{split}
\end{equation*}
where $\beta$ is the weight to tackle the class imbalance issue, and $d$ is the label indicating whether a token appears in the referential report. Finally, we prune the observation graph by masking out token-level nodes with $p(1|\bm{T}^A) < 0.5$ and masked token-level node representations denote as $\bm{T}^M=\text{Prune}(\bm{T})$.
\begin{figure}[t]
	\centering
    \setlength\belowcaptionskip{\fmargin}
    \includegraphics[width=1.0\linewidth]{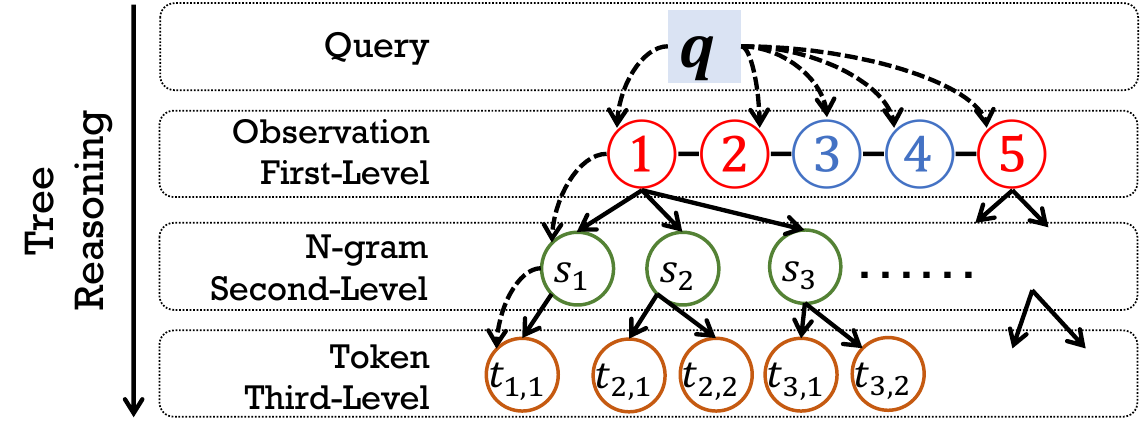}
	\caption{Illustration of the tree reasoning mechanism. It aggregates information from the observation level to the n-gram level and finally to the token level.}
    \label{figure: trr}
\end{figure}
\noindent\textbf{Tree Reasoning over Observation Graph.} We devise a tree reasoning (TrR) mechanism to aggregate observation-relevant information from the graph dynamically. The overall process is shown in Figure \ref{figure: trr}, where we aggregate node information from the observation level (i.e., first level) to the n-gram level (i.e., second level), then to the token level (i.e., third level). To be specific, given a query $\bm{q}^l$ and node representations at $l$-th level $\bm{k}^l \in \{\bm Z,\bm S,\bm T^M\}$, the tree reasoning path is $\bm{q}^0\xrightarrow{\bm{Z}}\bm{q}^1\xrightarrow{\bm{S}}\bm{q}^2\xrightarrow{\bm{T}^M}\bm{q}^3$, and the overall process, is formulated as below:
\begin{equation*}
\setlength{\belowdisplayskip}{\margin}
\setlength{\abovedisplayskip}{\margin}
    \begin{split}
        \bm{v}^{l+1}&=\text{MHA}(\bm{W}_q\bm{q}^{l},\bm{W}_k\bm{k}^{l},\bm{W}_v\bm{k}^{l}), \\
        \bm{q}^{l+1}&=\text{LayerNorm}(\bm{q}^{l}+\bm{v}^{l+1}), \\
    \end{split}
\end{equation*}
where MHA and LayerNorm are the multi-head self-attention, and layer normalization modules \cite{Transformer}, respectively. $\bm{W}_q$, $\bm{W}_k$, and $\bm{W}_v \in \mathbb{R}^{h\times h}$ are weight metrics for query, key, and value vector, respectively. Finally, we can obtain the multi-level information $\bm{q}^3$, containing observation, n-gram, and token information. 

\noindent\textbf{Report Generation with Tree Reasoning.} As shown in the right side of Figure \ref{figure: overall_framework}, an observation-guided Transformer decoder is devised to incorporate the graph information, including (i) multiple observation-guided decoder blocks (i.e., Decoder$_g$), which aims to align observations with the visual representations, and (ii) a tree-reasoning block (i.e., TrR$_g$), which aims to aggregate observation-relevant information. For Decoder$_g$, we insert an observation-related cross-attention module before a visually-aware cross-attention module. By doing this, the model can correctly focus on regions closely related to a specific observation. Given the visual representations $\bm h^v$, the node representations $\bm V=\{\bm Z,\bm S,\bm T^M\}$, and the hidden representation of the prefix $\bm{h}^w_* \in \mathbb{R}^h$, the $t$-th decoding step is formulated as:
\begin{equation*}
\setlength{\belowdisplayskip}{\margin}
\setlength{\abovedisplayskip}{\margin}
    \begin{split}
        \text{Decoder}_g&=\left\{
        \begin{aligned}
            \bm h^{s}_{t}&=\text{Self-Attn}(\bm h^{w}_{t},\bm h^{w}_{<t}, \bm h^{w}_{<t}), \\
            \bm h^{o}_{t}&=\text{Cross-Attn}(\bm h^{s}_t,\bm{Z},\bm{Z}), \\
            \bm h^{p}_t&=\text{Cross-Attn}(\bm h^{o}_{t},\bm h^v, \bm h^v), \\
        \end{aligned}
        \right. \\
        \text{TrR}_g&=\left\{
        \begin{aligned}
            \bm h^{d}_{t}&=\text{Self-Attn}(\bm h^{p}_{t},\bm h^{p}_{<t}, \bm h^{p}_{<t}), \\
            \bm q^3_t&=\text{TrR}(\bm h^{d}_t,[\bm Z,\bm S,\bm T^M]), \\
        \end{aligned}
        \right. \\
        p(y_t&|X,G,Y_{< t})=\text{Softmax}(\bm{W}_g\bm{q}^3_t + \bm{b}_g), \\
    \end{split}
\end{equation*}
where Self-Attn is the self-attention module, Cross-Attn is the cross-attention module, $\bm{h}^s_t, \bm{h}^o_t, \bm{h}^p_t \in \mathbb{R}^h$ are self-attended hidden state, observation-related hidden state, visually-aware hidden state of Decoder$_g$, respectively. $\bm{h}^d_t \in \mathbb{R}^h$ is the self-attended hidden state of TrR$_g$, $\bm{W}_g \in \mathbb{R}^{|V|\times h}$ is the weight matrix, and $\bm b_g \in \mathbb{R}^{|V|}$ is the bias vector. We omit other modules (i.e., Layer Normalization and Feed-Forward Network) in the standard Transformer for simplicity. Note that we extend the observation plan $Z$ to an observation graph $G$, so the probability of $y_t$ conditions on $G$ instead of $Z$. Then, we optimize the generation process using the negative log-likelihood loss:
\begin{equation*}
\setlength{\belowdisplayskip}{\margin}
\setlength{\abovedisplayskip}{\margin}
    \begin{split}
        \mathcal{L}_r=&-\sum^T_{t=1} \log p(y_t|X,G,Y_{< t}).
    \end{split}
\end{equation*}
Finally, the loss function of the generator is $\mathcal{L}_g=\mathcal{L}_r+\mathcal{L}_d$.
\begin{table*}[t]
    \centering
    \resizebox{\textwidth}{!}{
    \begin{tabular}{c|l|cccccc|ccc}
    \Xhline{2\arrayrulewidth}
    \multirow{2}{*}{\textbf{Dataset}} & \multirow{2}{*}{\textbf{Model}} & \multicolumn{6}{c|}{\textbf{NLG Metrics}} & \multicolumn{3}{c}{\textbf{CE Metrics}} \\
    \cline{3-11}
    & & \textbf{B-1} & \textbf{B-2} & \textbf{B-3} & \textbf{B-4} & \textbf{MTR} & \textbf{R-L} & \textbf{P} & \textbf{R} & \textbf{F}$_1$ \\ 
    \hline
    \multirow{10}{*}{\textsc{\makecell{IU \\ X-ray}}} & \textsc{R2Gen} & $0.470$ & $0.304$ & $0.219$ & $0.165$ & - & $0.371$ & - & - & - \\
    & \textsc{CA} & $0.492$ & $0.314$ & $0.222$ & $0.169$ & $0.193$ & $0.381$ & - & - & - \\
    & \textsc{CMCL} & $0.473$ & $0.305$ & $0.217$ & $0.162$ & $0.186$ & $0.378$ & - & - & - \\
    & \textsc{PPKED} & $0.483$ & $0.315$ & $0.224$ & $0.168$ & - & $0.376$ & - & - & - \\
    & \textsc{R2GenCMN} & $0.475$ & $0.309$ & $0.222$ & $0.170$ & $0.191$ & $0.375$ & - & - & - \\
    & $\mathcal{M}^2$\textsc{Tr} & $0.486$ & $0.317$ & $0.232$ & $0.173$ & $0.192$ & {$0.390$} & - & - & - \\
    & \textsc{AlignTransfomer} & $0.484$ & $0.313$ & $0.225$ & $0.173$ & - & $0.379$ & - & - & - \\
    & \textsc{KnowMat} & \underline{$0.496$} & {$0.327$} & {$0.238$} & $0.178$ & - & $0.381$ & - & - & - \\
    & \textsc{CMM-RL} & $0.494$ & $0.321$ & $0.235$ & {$0.181$} & {$0.201$} & {$0.384$} & - & - & - \\
    & \textsc{CMCA} & \underline{$0.496$} & \bm{$0.349$} & \bm{$0.268$} & \bm{$0.215$} & \bm{$0.209$} & \underline{$0.392$} & - & - & - \\
    & \textsc{ORGan} (Ours) & \bm{$0.510$} & \underline{$0.346$} & \underline{$0.255$} & \underline{$0.195$} & \underline{$0.205$} & \bm{$0.399$} & - & - & - \\
    \Xhline{2\arrayrulewidth}
    \multirow{10}{*}{\textsc{\makecell{MIMIC\\-CXR}}} & \textsc{R2Gen} & $0.353$ & $0.218$ & $0.145$ & $0.103$ & $0.142$ & $0.270$ & $0.333$ & $0.273$ & $0.276$ \\
    & \textsc{CA} & $0.350$ & $0.219$ & $0.152$ & $0.109$ & \underline{$0.151$} & $0.283$ & - & - & - \\
    & \textsc{CMCL} & $0.344$ & $0.217$ & $0.140$ & $0.097$ & $0.133$ & $0.281$ & - & - & - \\
    & \textsc{PPKED} & $0.360$ & $0.224$ & $0.149$ & $0.106$ & $0.149$ & $0.284$ & - & - & - \\
    & \textsc{R2GenCMN} & $0.353$ & $0.218$ & $0.148$ & $0.106$ & $0.142$ & $0.278$ & $0.344$ & $0.275$ & $0.278$ \\
    
    & $\mathcal{M}^2$\textsc{Tr} & $0.378$ & $0.232$ & $0.154$ & $0.107$ & $0.145$ & {$0.272$} & $0.240$ & \bm{$0.428$} & $0.308$ \\
    & \textsc{AlignTransfomer} & $0.378$ & \underline{$0.235$} & \underline{$0.156$} & $0.112$ & - & $0.283$ & - & - & - \\
    & \textsc{KnowMat} & $0.363$ & $0.228$ & \underline{$0.156$} & {$0.115$} & - & $0.284$ & \bm{$0.458$} & $0.348$ & $0.371$ \\
    & \textsc{CMM-RL} & \underline{$0.381$} & $0.232$ & $0.155$ & $0.109$ & \underline{$0.151$} & \underline{$0.287$} & $0.342$ & $0.294$ & $0.292$ \\
    & \textsc{CMCA} & $0.360$ & $0.227$ & \underline{$0.156$} & \underline{$0.117$} & $0.148$ & \underline{$0.287$} & \underline{$0.444$} & $0.297$ & $0.356$ \\
    & \textsc{ORGan} (Ours) & \bm{$0.407$} & \bm{$0.256$} & \bm{$0.172$} & \bm{$0.123$} & \bm{$0.162$} & \bm{$0.293$} & $0.416$ & \underline{$0.418$} & \bm{$0.385$}\\
    \Xhline{2\arrayrulewidth}
    \end{tabular}}
    \caption{Experimental Results of our model and baselines on the \textsc{IU X-ray} dataset and the \textsc{MIMIC-CXR} dataset. The best results are in \textbf{boldface}, and the \underline{underlined} are the second-best results.}
    \label{table: all_exps}
\end{table*}
\section{Experiments}
\subsection{Datasets}
Following previous research \cite{r2gen,r2gencmn}, we use two publicly available benchmarks to evaluate our method, which are \textsc{IU X-ray}\footnote{\url{https://openi.nlm.nih.gov/}} \cite{iu_xray} and \textsc{MIMIC-CXR}\footnote{\url{https://physionet.org/content/MIMIC-cxr-jpg/2.0.0/}} \cite{mimic_cxr}. Both datasets have been automatically de-identified, and we use the same preprocessing setup of \citet{r2gen}.
\begin{itemize}[noitemsep,topsep=2pt]
    \item \textsc{IU X-ray} is collected by Indiana University, containing 3,955 reports with two X-ray images per report resulting in 7,470 images in total. We split the dataset into train/validation/test sets with a ratio of 7:1:2, which is the same data split as in \cite{r2gen}.
    \item \textsc{MIMIC-CXR} consists of 377,110 chest X-ray images and 227,827 reports from 63,478 patients. We adopt the standard train/validation/test splits.
\end{itemize}

\subsection{Evaluation Metrics and Baselines}
We adopt natural language generation metrics (NLG Metrics) and clinical efficacy (CE Metrics) to evaluate the models. BLEU \cite{bleu}, METEOR \cite{meteor}, and ROUGE \cite{rouge} are selected as NLG Metrics, and we use the MS-COCO caption evaluation tool\footnote{\url{https://github.com/tylin/coco-caption}} to compute the results. For CE Metrics, we adopt CheXpert \cite{chexpert} for \textsc{MIMIC-CXR} dataset to label the generated reports compared with disease labels of the references. 

To evaluate the performance of \textsc{ORGan}, we compare it with the following 10 state-of-the-art (SOTA) baselines: \textsc{R2Gen} \cite{r2gen}, \textsc{CA} \cite{ca}, \textsc{CMCL} \cite{cmcl}, \textsc{PPKED} \cite{ppked}, \textsc{R2GenCMN} \cite{r2gencmn}, \textsc{AlignTransformer} \cite{aligntransformer}, \textsc{KnowMat} \cite{mia}, $\mathcal{M}^2$\textsc{Tr} \cite{m2tr}, \textsc{CMM-RL} \cite{cmm-rl}, and \textsc{CMCA} \cite{cmca}. 
\begin{table*}[t]
    \centering
    \resizebox{\textwidth}{!}{
    \begin{tabular}{c|l|cccccc|ccc}
    \Xhline{2\arrayrulewidth}
    \multirow{2}{*}{\textbf{Dataset}} & \multirow{2}{*}{\textbf{Model}} & \multicolumn{6}{c|}{\textbf{NLG Metrics}} & \multicolumn{3}{c}{\textbf{CE Metrics}} \\\cline{3-11}
    & & \textbf{B-1} & \textbf{B-2} & \textbf{B-3} & \textbf{B-4} & \textbf{MTR} & \textbf{R-L} & \textbf{P} & \textbf{R} & \textbf{F}$_1$ \\ 
    \hline
    \multirow{4}{*}{\textsc{\makecell{IU \\ X-ray}}} & \textsc{ORGan} & {$0.510$} & {$0.346$} & {$0.255$} & {$0.195$} & {$0.205$} & {$0.399$} & - & - & - \\
    & \textsc{ORGan} \textit{w/o} Plan & $0.406$ & $0.254$ & $0.178$ & $0.133$ & $0.167$ & $0.372$ & - & - & - \\
    & \textsc{ORGan} \textit{w/o} Graph & $0.461$ & $0.302$ & $0.218$ & $0.164$ & $0.186$ & $0.383$  & - & - & - \\
    & \textsc{ORGan} \textit{w/o} TrR & $0.494$ & $0.335$ & $0.247$ & $0.190$ & $0.203$ & $0.395$ & - & - & - \\
    \Xhline{2\arrayrulewidth}
    \multirow{4}{*}{\textsc{\makecell{MIMIC\\-CXR}}} & \textsc{ORGan} & {$0.407$} & {$0.256$} & {$0.172$} & {$0.123$} & {$0.162$} & {$0.293$} & {$0.416$} & {$0.418$} & {$0.385$} \\
    & \textsc{ORGan} \textit{w/o} Plan & $0.334$ & $0.211$ & $0.145$ & $0.107$ & $0.136$ & $0.282$ & $0.384$ & $0.239$ & $0.252$ \\
    & \textsc{ORGan} \textit{w/o} Graph & $0.369$ & $0.233$ & $0.158$ & $0.113$ & $0.151$ & $0.290$ &  $0.401$ & $0.415$ & $0.383$\\
    & \textsc{ORGan} \textit{w/o} TrR & $0.405$ & $0.254$ & $0.170$ & $0.121$ & $0.161$ & $0.291$ & $0.411$ & $0.419$ & $0.386$ \\
    \Xhline{2\arrayrulewidth}
    \end{tabular}}
    \caption{Ablation results of our model and its variants, where \textsc{ORGan} \textit{w/o} Plan is the standard Transformer model.}
    \label{table: ablation}
\end{table*}
\subsection{Implementation Details}
We adopt the ResNet-101 \cite{resnet} pretrained on ImageNet \cite{imagenet} as the visual extractor. For \textsc{IU X-ray}, we further fine-tune ResNet-101 on CheXpert \cite{chexpert}. The layer number of all the encoders and decoders is set to 3 except for Graph Encoder, where the layer number is set to 2. The input dimension and the feed-forward network dimension of a Transformer block are set to 512, and each block contains 8 attention heads. The beam size for decoding is set to 4, and the maximum decoding step is set to 64/104 for \textsc{IU X-ray} and \textsc{MIMIC-CXR}, respectively.

We use AdamW \cite{adamw} as the optimizer and set the initial learning rate for the visual extractor as 5e-5 and 1e-4 for the rest of the parameters, with a linear schedule decreasing from the initial learning rate to 0. $\alpha$ is set to 0.5, the dropout rate is set to 0.1, and the batch size is set to 32. For IU X-ray, we train the planner/generator for 15/15 epochs, and $\beta$ is set to 2. For MIMIC-CXR, the training epoch of the planner/generator is set to 3/5, and $\beta$ is set to 5. We select the best checkpoints of the planner based on micro F$_1$ of all observations and select the generator based on the BLEU-4 on the validation set. Our model has 65.9M parameters, and the implementations are based on HuggingFace's Transformers \cite{huggingface}. We conduct all the experiments on an NVIDIA-3090 GTX GPU with mixed precision. The NLTK package version is 3.6.2.

\section{Results}
\subsection{NLG Results}
Table \ref{table: all_exps} shows the experimental results. \textsc{ORGan} outperforms most of the baselines (except CMCA on \textsc{IU X-ray}) and achieves state-of-the-art performance. Specifically, our model achieves $0.195$ BLEU-4 on the \textsc{IU X-xray} dataset, which is the second-best result, and $0.123$ BLEU-4 on the \textsc{MIMIC-CXR} dataset, leading to a $5.1\%$ increment of compared to the best baseline (i.e., CMCA). In terms of METEOR, \textsc{ORGan} achieves competitive performance on both datasets. In addition, our model increases R-L by $0.6\%$ on the \textsc{MIMIC-CXR} dataset compared to the best baseline and achieves the second-best result on the \textsc{IU X-ray} dataset. This indicates that by introducing the guidance of observations, \textsc{ORGan} can generate more coherent text than baselines. However, we notice that on the \textsc{IU X-ray} dataset, there is still a performance gap between our model and the best baseline (i.e., CMCA). The reason may be that the overall data size of this dataset is small ($\sim$ 2,000 samples for training). It is difficult to train a good planner using a small training set, especially with cross-modal data. As we can see from Table \ref{table: planner_exp}, the planner only achieves $0.132$ Macro-F$_1$ on the \textsc{IU X-ray} dataset, which is relatively low compared to the performance of the planner on the \textsc{MIMIC-CXR} dataset. Thus, accumulation errors unavoidably propagate to the generator, which leads to lower performance.
\begin{table}[t]
    \centering
    \small
    {
    \begin{tabular}{c|ccc}
    \Xhline{2\arrayrulewidth}
    \textbf{Dataset} & \textbf{Micro-F}$_1$ & \textbf{Macro-F}$_1$ & \textbf{B-2} \\ \hline
    \textsc{IU X-ray} & $0.507$ & $0.132$ & $0.499$\\\hline
    \textsc{MIMIC-CXR} & $0.574$ & $0.397$ & $0.357$ \\
    \Xhline{2\arrayrulewidth}
    \end{tabular}}
    \caption{Experimental results of observation planning. Macro-F$_1$ and Micro-F$_1$ denote the macro F$_1$ and micro F$_1$ of abnormal observations, respectively.}
    \label{table: planner_exp}
\end{table}
\begin{table}[t]
    \centering
    \small
    {
    \begin{tabular}{c|c|ccc}
        \Xhline{2\arrayrulewidth}
        \textbf{Dataset} &\textbf{K} & \textbf{B-2/4} & \textbf{MTR} & \textbf{R-L} \\ \hline
        \multirow{3}{*}{\textsc{\makecell{IU\\X-ray}}} &$10$ & $0.309$/$0.170$ & $0.192$ & $0.388$\\
        & $20$ & $0.333$/$0.180$ & $0.202$ & $0.393$  \\
        & $30$ & $0.346$/$0.195$ & $0.205$ & $0.399$ \\
        \Xhline{2\arrayrulewidth}
        \multirow{3}{*}{\textsc{\makecell{MIMIC\\-CXR}}} &$10$ & $0.249$/$0.118$ & $0.161$ & $0.290$ \\
        & $20$ & $0.252$/$0.120$ & $0.159$ & $0.292$\\ 
        & $30$ & $0.256$/$0.123$  & $0.162$ & $0.293$ \\
        \Xhline{2\arrayrulewidth}
    \end{tabular}}
    \caption{Experimental results under the different number (K) of selected n-grams.}
    \label{table: k_exp}
\end{table}

\subsection{Clinical Efficacy Results}
The clinical efficacy results are listed on the right side of Table \ref{table: all_exps}. On the \textsc{MIMIC-CXR} dataset, our model outperforms previous SOTA results. Specifically, our model achieves $0.385$ F$_1$, increasing by $1.4\%$ compared to the best baseline. In addition, $0.416$ precision and $0.418$ recall are achieved by \textsc{ORGan}, which are competitive results. This indicates that our model can successfully maintain the clinical consistency between the images and the reports.
\begin{figure*}[t]
	\centering
    \setlength\belowcaptionskip{\fmargin}
    \includegraphics[width=1.0\linewidth]{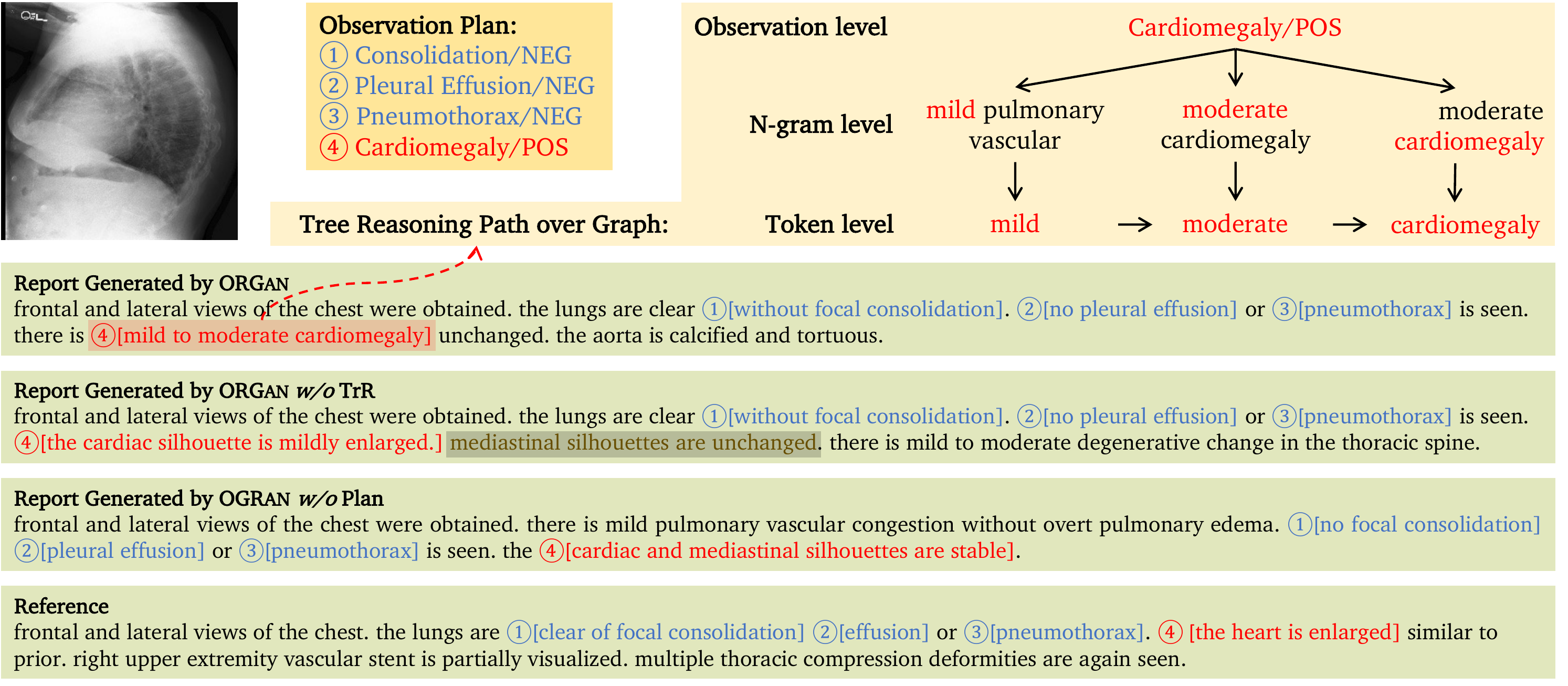}
	\caption{Case study of our model with the tree reasoning path of the mention "\textit{mild to moderate cardiomegaly}."}
    \label{figure: case_study}
\end{figure*}
\begin{figure}[t]
	\centering
    \setlength\belowcaptionskip{\fmargin}
    \includegraphics[width=1.0\linewidth]{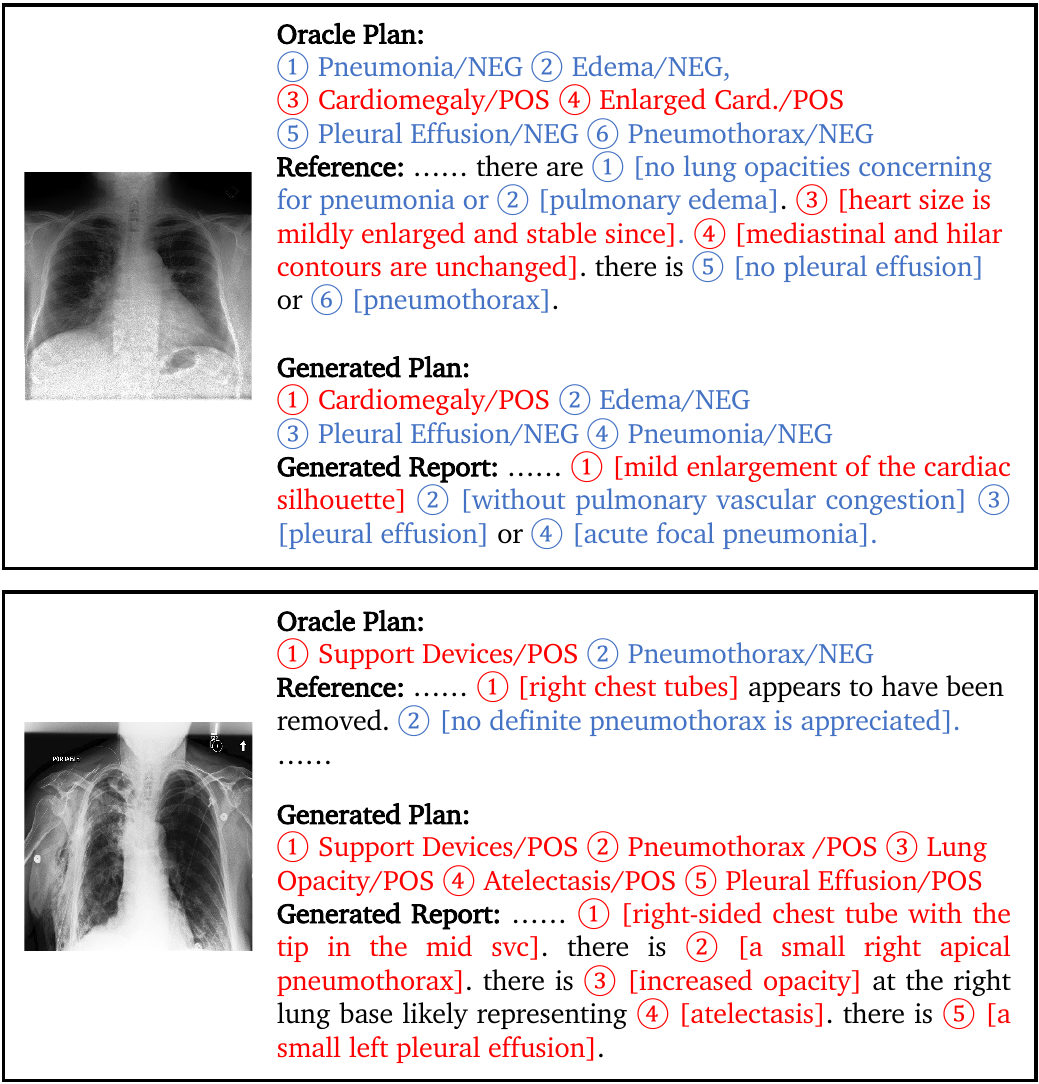}
	\caption{Examples of error cases. \textit{Enlarged Card.} refers to \textit{Enlarged Cardiomediastinum}. The upper case omits one positive observation and the bottom case contains false positive observations.}
    \label{figure: failure_cases}
\end{figure}
\subsection{Ablation Results}
To examine the effect of the observation plan and the TrR mechanism, we perform ablation tests, and the ablation results are listed in Table \ref{table: ablation}. There are three variants: (1) \textsc{ORGan} \textit{w/o} Plan, which does not consider observation information, (2) \textsc{ORGan} \textit{w/o} Graph, which only considers observations but not the observation graph, (3) \textsc{ORGan} \textit{w/o} TrR, which select information without using the TrR mechanism. Compared to the full model, the performance of \textsc{ORGan} \textit{w/o} Plan drops significantly on both datasets. This indicates that observation information plays a vital role in generating reports. For \textsc{ORGan} \textit{w/o} Graph, the performance on NLG metrics decreases significantly, but the performance of clinical efficacy remains nearly the same as the full model. This is reasonable because the observation graph is designed to enrich the observation plan to achieve better word-level realization. On the performance of \textsc{ORGan} \textit{w/o} TrR, a similar result of \textsc{ORGan} \textit{w/o} Graph is observed. This indicates that TrR can enrich the plan information, and stronger reasoning can help generate high-quality reports.

We also conduct experiments on the impact of the number (K) of selected n-grams, as shown in Table \ref{table: k_exp}. There is a performance gain when increasing K from $10$ to $20$ and to $30$ on both datasets. On the \textsc{IU X-ray} dataset, B-2 increases by $2.4\%$ and $3.7\%$ and B-4 rises by $1.0\%$ and $1.5\%$. A similar trend is also observed on the \textsc{MIMIC-CXR} dataset.

\subsection{Qualitative Analysis}
We conduct a case study and analyze some error cases on the \textsc{MIMIC-CXR} dataset to provide more insights.

\noindent\textbf{Case Study.} We conduct a case study to show how the observation and the tree reasoning mechanism guide the report generation process, as shown in Figure \ref{figure: case_study}. We show the generated reports of \textsc{ORGan}, \textsc{ORGan} \textit{w/o} TrR, and \textsc{ORGan} \textit{w/o} Plan, respectively. All three models successfully generate the first three negative observations and the last positive observation. However, variant \textit{w/o} plan generates "\textit{mild pulmonary vascular congestion without overt pulmonary edema}" which is not consistent with the radiograph. In terms of the output of variant \textit{w/o} TrR, "\textit{mediastinal silhouettes are unchanged}" is closely related to observation \textit{Enlarged Cardiomediastinum} instead of \textit{Cardiomegaly}. Only \textsc{ORGan} can generate the \textit{Cardiomegaly}/POS presented in the observation plan with a TrR path. This indicates that observations play a vital role in maintaining clinical consistency. In addition, most of the tokens in the observation mention \textit{mild to moderate cardiomegaly} can be found in the observation graph, which demonstrates that the graph can provide useful information in word-level realization.

\noindent\textbf{Error Analysis.} We depicte error cases generated by \textsc{ORGan} in Figure \ref{figure: failure_cases}. The major error is caused by introducing incorrect observation plans. Specifically, the generated plan of the upper case omits one positive observation (i.e., \textit{Enlarged Cardiomediastinum}/POS), resulting in false negative observations in its corresponding generated report. Another error is false positive observations appearing in the generated reports (e.g., the bottom case). Thus, how to improve the performance of the planner is a potential future work to enhance clinical accuracy.
\section{Related Work}
\subsection{Image Captioning and Medical Report Generation}
Image Captioning \cite{ST,att2in,adaatt,topdown} has long been an attractive research topic, and there has been a surging interest in developing medical AI applications. Medical Report Generation \cite{coatt,hrgr} is one of these applications. \citet{r2gen} proposed a memory-driven Transformer model to generate radiology reports. \citet{r2gencmn} further proposed a cross-modal memory network to facilitate report generation. \citet{cmm-rl} proposed to utilize reinforcement learning \cite{reinforce} to align the cross-modal information between the image and the corresponding report. In addition to these methods, \citet{ca} proposed the Contrastive Attention model comparing the given image with normal images to distill information. \citet{mia} proposed to introduce general and specific knowledge extracted from RadGraph \cite{radgraph} in report generation. \citet{cmcl} proposed a competence-based multimodal curriculum learning to guide the learning process. \citet{ppked} proposed to explore and distill posterior and prior knowledge for report generation.

Several research works focus on improving the clinical accuracy of the generated reports. \citet{clinical_reward} proposed a clinically coherent reward for clinically accurate reinforcement learning to improve clinical accuracy. \citet{clinically_coherent} proposed to use CheXpert \cite{chexpert} as a source of clinical information to generate clinically coherent reports. \citet{fact_ent} proposed to use entity matching score as a reward to encourage the model to generate factually complete and consistent radiology reports. \citet{coplan} proposed a planning-based method and regarded the report generation task as the data-to-text generation task.

\subsection{Planning in Text Generation}
Another line of research closely related to our work is planning in text generation, which has been applied to multiple tasks (e.g., Data-to-Text Generation, Summarization, and Story Generation). \citet{pair} propose a global planning and iterative refinement model for long text generation. \citet{ssplaner} propose a self-supervised planning framework for paragraph completion. \citet{planet} propose a dynamic planning model for long-form text generation to tackle the issue of incoherence outputs. \citet{step_by_step} proposed a neural data-to-text generation by separating planning from realization. \citet{plan_then_generate} proposed a controlled data-to-text generation framework by planning the order of content in a table.

\section{Conclusion}
In this paper, we propose \textsc{ORGan}, an observation-guided radiology report generation framework, which first produces an observation plan and then generates the corresponding report based on the radiograph and the plan. To achieve better observation realization, we construct a three-level observation graph containing observations, observation-aware n-grams, and tokens, and we propose a tree reasoning mechanism to capture observation-related information by dynamically aggregating nodes in the graph. Experimental results demonstrate the effectiveness of our proposed framework in terms of maintaining the clinical consistency between radiographs and generated reports. 

\section*{Limitations}
There are several limitations to our framework. Specifically, since observations are introduced as guiding information, our framework requires observation extraction tools to label the training set in advance.  Then, the nodes contained in the observation graph are mined from the training data. As a result, the mined n-grams could be biased when the overall size of the training set is small. In addition, our framework is a pipeline, and the report generation performance highly relies on the performance of observation planning. Thus, errors could accumulate through the pipeline, especially for small datasets. Finally, our framework is designed for radiology report generation targeting Chest X-ray images. However, there are other types of medical images (e.g., Fundus Fluorescein Angiography images) that our framework needs to examine.
\section*{Ethics Statement}
The \textsc{IU X-ray}\cite{iu_xray} and \textsc{MIMIC-CXR} \cite{mimic_cxr} datasets have been automatically de-identified to protect patient privacy. The proposed system is intended to generate radiology reports automatically, alleviating the workload of radiologists. However, we notice that the proposed system can generate false positive observations and inaccurate diagnoses due to systematic biases. If the system, as deployed, would learn from further user input (i.e., patients' radiographs), there are risks of personal information leakage while interacting with the system. This might be mitigated by using anonymous technology to protect privacy. Thus, we urge users to cautiously examine the ethical implications of the generated output in real-world applications.
\section*{Acknolwedgments}
This work was supported in part by General Program of National Natural Science Foundation of China (Grant No. 82272086, 62076212), Guangdong Provincial Department of Education (Grant No. 2020ZDZX3043), Shenzhen Natural Science Fund (JCYJ20200109140820699 and the Stable Support Plan Program 20200925174052004), and the Research Grants Council of Hong Kong (15207920, 15207821, 15207122).

\bibliography{acl_latex}

\begin{thebibliography}{42}
\expandafter\ifx\csname natexlab\endcsname\relax\def\natexlab#1{#1}\fi

\bibitem[{Anderson et~al.(2018)Anderson, He, Buehler, Teney, Johnson, Gould,
  and Zhang}]{topdown}
Peter Anderson, Xiaodong He, Chris Buehler, Damien Teney, Mark Johnson, Stephen
  Gould, and Lei Zhang. 2018.
\newblock \href {https://doi.org/10.1109/CVPR.2018.00636} {Bottom-up and
  top-down attention for image captioning and visual question answering}.
\newblock In \emph{2018 {IEEE} Conference on Computer Vision and Pattern
  Recognition, {CVPR} 2018, Salt Lake City, UT, USA, June 18-22, 2018}, pages
  6077--6086. Computer Vision Foundation / {IEEE} Computer Society.

\bibitem[{Banerjee and Lavie(2005)}]{meteor}
Satanjeev Banerjee and Alon Lavie. 2005.
\newblock \href {https://aclanthology.org/W05-0909} {{METEOR}: An automatic
  metric for {MT} evaluation with improved correlation with human judgments}.
\newblock In \emph{Proceedings of the {ACL} Workshop on Intrinsic and Extrinsic
  Evaluation Measures for Machine Translation and/or Summarization}, pages
  65--72, Ann Arbor, Michigan. Association for Computational Linguistics.

\bibitem[{Chen et~al.(2021)Chen, Shen, Song, and Wan}]{r2gencmn}
Zhihong Chen, Yaling Shen, Yan Song, and Xiang Wan. 2021.
\newblock \href {https://doi.org/10.18653/v1/2021.acl-long.459} {Cross-modal
  memory networks for radiology report generation}.
\newblock In \emph{Proceedings of the 59th Annual Meeting of the Association
  for Computational Linguistics and the 11th International Joint Conference on
  Natural Language Processing, {ACL/IJCNLP} 2021, (Volume 1: Long Papers),
  Virtual Event, August 1-6, 2021}, pages 5904--5914. Association for
  Computational Linguistics.

\bibitem[{Chen et~al.(2020)Chen, Song, Chang, and Wan}]{r2gen}
Zhihong Chen, Yan Song, Tsung-Hui Chang, and Xiang Wan. 2020.
\newblock Generating radiology reports via memory-driven transformer.
\newblock In \emph{Proceedings of the 2020 Conference on Empirical Methods in
  Natural Language Processing}.

\bibitem[{Church and Hanks(1990)}]{pmi}
Kenneth~Ward Church and Patrick Hanks. 1990.
\newblock \href {https://aclanthology.org/J90-1003} {Word association norms,
  mutual information, and lexicography}.
\newblock \emph{Computational Linguistics}, 16(1):22--29.

\bibitem[{Demner-Fushman et~al.(2016)Demner-Fushman, Kohli, Rosenman, Shooshan,
  Rodriguez, Antani, Thoma, and McDonald}]{iu_xray}
Dina Demner-Fushman, Marc~D Kohli, Marc~B Rosenman, Sonya~E Shooshan, Laritza
  Rodriguez, Sameer Antani, George~R Thoma, and Clement~J McDonald. 2016.
\newblock Preparing a collection of radiology examinations for distribution and
  retrieval.
\newblock \emph{Journal of the American Medical Informatics Association},
  23(2):304--310.

\bibitem[{Deng et~al.(2009)Deng, Dong, Socher, Li, Li, and Fei-Fei}]{imagenet}
Jia Deng, Wei Dong, Richard Socher, Li-Jia Li, Kai Li, and Li~Fei-Fei. 2009.
\newblock \href {https://doi.org/10.1109/CVPR.2009.5206848} {Imagenet: A
  large-scale hierarchical image database}.
\newblock In \emph{2009 IEEE Conference on Computer Vision and Pattern
  Recognition}, pages 248--255.

\bibitem[{Diao et~al.(2021)Diao, Xu, Su, Jiang, Song, and Zhang}]{ngram_pmi}
Shizhe Diao, Ruijia Xu, Hongjin Su, Yilei Jiang, Yan Song, and Tong Zhang.
  2021.
\newblock \href {https://doi.org/10.18653/v1/2021.acl-long.259} {Taming
  pre-trained language models with n-gram representations for low-resource
  domain adaptation}.
\newblock In \emph{Proceedings of the 59th Annual Meeting of the Association
  for Computational Linguistics and the 11th International Joint Conference on
  Natural Language Processing (Volume 1: Long Papers)}, pages 3336--3349,
  Online. Association for Computational Linguistics.

\bibitem[{He et~al.(2015)He, Zhang, Ren, and Sun}]{resnet}
Kaiming He, Xiangyu Zhang, Shaoqing Ren, and Jian Sun. 2015.
\newblock Deep residual learning for image recognition.
\newblock \emph{arXiv preprint arXiv:1512.03385}.

\bibitem[{Hu et~al.(2022)Hu, Chan, Liu, Xiao, Wu, and Huang}]{planet}
Zhe Hu, Hou~Pong Chan, Jiachen Liu, Xinyan Xiao, Hua Wu, and Lifu Huang. 2022.
\newblock \href {https://doi.org/10.18653/v1/2022.acl-long.163} {{PLANET}:
  Dynamic content planning in autoregressive transformers for long-form text
  generation}.
\newblock In \emph{Proceedings of the 60th Annual Meeting of the Association
  for Computational Linguistics (Volume 1: Long Papers)}, pages 2288--2305,
  Dublin, Ireland. Association for Computational Linguistics.

\bibitem[{Hua and Wang(2020)}]{pair}
Xinyu Hua and Lu~Wang. 2020.
\newblock \href {https://doi.org/10.18653/v1/2020.emnlp-main.57} {{PAIR}:
  Planning and iterative refinement in pre-trained transformers for long text
  generation}.
\newblock In \emph{Proceedings of the 2020 Conference on Empirical Methods in
  Natural Language Processing (EMNLP)}, pages 781--793, Online. Association for
  Computational Linguistics.

\bibitem[{Irvin et~al.(2019)Irvin, Rajpurkar, Ko, Yu, Ciurea{-}Ilcus, Chute,
  Marklund, Haghgoo, Ball, Shpanskaya, Seekins, Mong, Halabi, Sandberg, Jones,
  Larson, Langlotz, Patel, Lungren, and Ng}]{chexpert}
Jeremy Irvin, Pranav Rajpurkar, Michael Ko, Yifan Yu, Silviana Ciurea{-}Ilcus,
  Chris Chute, Henrik Marklund, Behzad Haghgoo, Robyn~L. Ball, Katie~S.
  Shpanskaya, Jayne Seekins, David~A. Mong, Safwan~S. Halabi, Jesse~K.
  Sandberg, Ricky Jones, David~B. Larson, Curtis~P. Langlotz, Bhavik~N. Patel,
  Matthew~P. Lungren, and Andrew~Y. Ng. 2019.
\newblock \href {https://doi.org/10.1609/aaai.v33i01.3301590} {Chexpert: {A}
  large chest radiograph dataset with uncertainty labels and expert
  comparison}.
\newblock In \emph{The Thirty-Third {AAAI} Conference on Artificial
  Intelligence, {AAAI} 2019, The Thirty-First Innovative Applications of
  Artificial Intelligence Conference, {IAAI} 2019, The Ninth {AAAI} Symposium
  on Educational Advances in Artificial Intelligence, {EAAI} 2019, Honolulu,
  Hawaii, USA, January 27 - February 1, 2019}, pages 590--597. {AAAI} Press.

\bibitem[{Jain et~al.(2021)Jain, Agrawal, Saporta, Truong, Duong, Bui, Chambon,
  Zhang, Lungren, Ng, Langlotz, and Rajpurkar}]{radgraph}
Saahil Jain, Ashwin Agrawal, Adriel Saporta, Steven Q.~H. Truong, Du~Nguyen
  Duong, Tan Bui, Pierre Chambon, Yuhao Zhang, Matthew~P. Lungren, Andrew~Y.
  Ng, Curtis~P. Langlotz, and Pranav Rajpurkar. 2021.
\newblock \href {http://arxiv.org/abs/2106.14463} {Radgraph: Extracting
  clinical entities and relations from radiology reports}.
\newblock \emph{CoRR}, abs/2106.14463.

\bibitem[{Jing et~al.(2018)Jing, Xie, and Xing}]{coatt}
Baoyu Jing, Pengtao Xie, and Eric~P. Xing. 2018.
\newblock \href {https://doi.org/10.18653/v1/P18-1240} {On the automatic
  generation of medical imaging reports}.
\newblock In \emph{Proceedings of the 56th Annual Meeting of the Association
  for Computational Linguistics, {ACL} 2018, Melbourne, Australia, July 15-20,
  2018, Volume 1: Long Papers}, pages 2577--2586. Association for Computational
  Linguistics.

\bibitem[{Johnson et~al.(2019)Johnson, Pollard, Greenbaum, Lungren, Deng, Peng,
  Lu, Mark, Berkowitz, and Horng}]{mimic_cxr}
Alistair~EW Johnson, Tom~J Pollard, Nathaniel~R Greenbaum, Matthew~P Lungren,
  Chih-ying Deng, Yifan Peng, Zhiyong Lu, Roger~G Mark, Seth~J Berkowitz, and
  Steven Horng. 2019.
\newblock Mimic-cxr-jpg, a large publicly available database of labeled chest
  radiographs.
\newblock \emph{arXiv preprint arXiv:1901.07042}.

\bibitem[{Kang and Hovy(2020)}]{ssplaner}
Dongyeop Kang and Eduard Hovy. 2020.
\newblock \href {https://doi.org/10.18653/v1/2020.emnlp-main.529} {Plan ahead:
  Self-supervised text planning for paragraph completion task}.
\newblock In \emph{Proceedings of the 2020 Conference on Empirical Methods in
  Natural Language Processing (EMNLP)}, pages 6533--6543, Online. Association
  for Computational Linguistics.

\bibitem[{Li et~al.(2018)Li, Liang, Hu, and Xing}]{hrgr}
Yuan Li, Xiaodan Liang, Zhiting Hu, and Eric~P. Xing. 2018.
\newblock \href
  {https://proceedings.neurips.cc/paper/2018/hash/e07413354875be01a996dc560274708e-Abstract.html}
  {Hybrid retrieval-generation reinforced agent for medical image report
  generation}.
\newblock In \emph{Advances in Neural Information Processing Systems 31: Annual
  Conference on Neural Information Processing Systems 2018, NeurIPS 2018,
  December 3-8, 2018, Montr{\'{e}}al, Canada}, pages 1537--1547.

\bibitem[{Lin(2004)}]{rouge}
Chin-Yew Lin. 2004.
\newblock \href {https://aclanthology.org/W04-1013} {{ROUGE}: A package for
  automatic evaluation of summaries}.
\newblock In \emph{Text Summarization Branches Out}, pages 74--81, Barcelona,
  Spain. Association for Computational Linguistics.

\bibitem[{Liu et~al.(2021{\natexlab{a}})Liu, Ge, and Wu}]{cmcl}
Fenglin Liu, Shen Ge, and Xian Wu. 2021{\natexlab{a}}.
\newblock \href {https://doi.org/10.18653/v1/2021.acl-long.234}
  {Competence-based multimodal curriculum learning for medical report
  generation}.
\newblock In \emph{Proceedings of the 59th Annual Meeting of the Association
  for Computational Linguistics and the 11th International Joint Conference on
  Natural Language Processing, {ACL/IJCNLP} 2021, (Volume 1: Long Papers),
  Virtual Event, August 1-6, 2021}, pages 3001--3012. Association for
  Computational Linguistics.

\bibitem[{Liu et~al.(2021{\natexlab{b}})Liu, Wu, Ge, Fan, and Zou}]{ppked}
Fenglin Liu, Xian Wu, Shen Ge, Wei Fan, and Yuexian Zou. 2021{\natexlab{b}}.
\newblock \href
  {https://openaccess.thecvf.com/content/CVPR2021/html/Liu\_Exploring\_and\_Distilling\_Posterior\_and\_Prior\_Knowledge\_for\_Radiology\_Report\_CVPR\_2021\_paper.html}
  {Exploring and distilling posterior and prior knowledge for radiology report
  generation}.
\newblock In \emph{{IEEE} Conference on Computer Vision and Pattern
  Recognition, {CVPR} 2021, virtual, June 19-25, 2021}, pages 13753--13762.
  Computer Vision Foundation / {IEEE}.

\bibitem[{Liu et~al.(2021{\natexlab{c}})Liu, Yin, Wu, Ge, Zhang, and Sun}]{ca}
Fenglin Liu, Changchang Yin, Xian Wu, Shen Ge, Ping Zhang, and Xu~Sun.
  2021{\natexlab{c}}.
\newblock \href {https://doi.org/10.18653/v1/2021.findings-acl.23} {Contrastive
  attention for automatic chest x-ray report generation}.
\newblock In \emph{Findings of the Association for Computational Linguistics:
  {ACL/IJCNLP} 2021, Online Event, August 1-6, 2021}, volume {ACL/IJCNLP} 2021
  of \emph{Findings of {ACL}}, pages 269--280. Association for Computational
  Linguistics.

\bibitem[{Liu et~al.(2019)Liu, Hsu, McDermott, Boag, Weng, Szolovits, and
  Ghassemi}]{clinical_reward}
Guanxiong Liu, Tzu{-}Ming~Harry Hsu, Matthew B.~A. McDermott, Willie Boag,
  Wei{-}Hung Weng, Peter Szolovits, and Marzyeh Ghassemi. 2019.
\newblock \href {http://arxiv.org/abs/1904.02633} {Clinically accurate chest
  x-ray report generation}.
\newblock \emph{CoRR}, abs/1904.02633.

\bibitem[{Loshchilov and Hutter(2019)}]{adamw}
Ilya Loshchilov and Frank Hutter. 2019.
\newblock \href {https://openreview.net/forum?id=Bkg6RiCqY7} {Decoupled weight
  decay regularization}.
\newblock In \emph{7th International Conference on Learning Representations,
  {ICLR} 2019, New Orleans, LA, USA, May 6-9, 2019}. OpenReview.net.

\bibitem[{Lovelace and Mortazavi(2020)}]{clinically_coherent}
Justin Lovelace and Bobak Mortazavi. 2020.
\newblock \href {https://doi.org/10.18653/v1/2020.findings-emnlp.110} {Learning
  to generate clinically coherent chest {X}-ray reports}.
\newblock In \emph{Findings of the Association for Computational Linguistics:
  EMNLP 2020}, pages 1235--1243, Online. Association for Computational
  Linguistics.

\bibitem[{Lu et~al.(2017)Lu, Xiong, Parikh, and Socher}]{adaatt}
Jiasen Lu, Caiming Xiong, Devi Parikh, and Richard Socher. 2017.
\newblock \href {https://doi.org/10.1109/CVPR.2017.345} {Knowing when to look:
  Adaptive attention via a visual sentinel for image captioning}.
\newblock In \emph{2017 {IEEE} Conference on Computer Vision and Pattern
  Recognition, {CVPR} 2017, Honolulu, HI, USA, July 21-26, 2017}, pages
  3242--3250. {IEEE} Computer Society.

\bibitem[{Miura et~al.(2021)Miura, Zhang, Tsai, Langlotz, and
  Jurafsky}]{fact_ent}
Yasuhide Miura, Yuhao Zhang, Emily Tsai, Curtis Langlotz, and Dan Jurafsky.
  2021.
\newblock \href {https://doi.org/10.18653/v1/2021.naacl-main.416} {Improving
  factual completeness and consistency of image-to-text radiology report
  generation}.
\newblock In \emph{Proceedings of the 2021 Conference of the North American
  Chapter of the Association for Computational Linguistics: Human Language
  Technologies}, pages 5288--5304, Online. Association for Computational
  Linguistics.

\bibitem[{Moryossef et~al.(2019)Moryossef, Goldberg, and Dagan}]{step_by_step}
Amit Moryossef, Yoav Goldberg, and Ido Dagan. 2019.
\newblock \href {https://doi.org/10.18653/v1/N19-1236} {{S}tep-by-step:
  {S}eparating planning from realization in neural data-to-text generation}.
\newblock In \emph{Proceedings of the 2019 Conference of the North {A}merican
  Chapter of the Association for Computational Linguistics: Human Language
  Technologies, Volume 1 (Long and Short Papers)}, pages 2267--2277,
  Minneapolis, Minnesota. Association for Computational Linguistics.

\bibitem[{Nishino et~al.(2022)Nishino, Miura, Taniguchi, Ohkuma, Suzuki, Kido,
  and Tomiyama}]{coplan}
Toru Nishino, Yasuhide Miura, Tomoki Taniguchi, Tomoko Ohkuma, Yuki Suzuki,
  Shoji Kido, and Noriyuki Tomiyama. 2022.
\newblock \href
  {https://preview.aclanthology.org/emnlp-22-ingestion/2022.emnlp-main.480/}
  {Factual accuracy is not enough: Planning consistent description order for
  radiology report generation}.
\newblock In \emph{Proceedings of the 2022 Conference on Empirical Methods in
  Natural Language Processing}, Online. Association for Computational
  Linguistics.

\bibitem[{Nooralahzadeh et~al.(2021)Nooralahzadeh, Perez~Gonzalez,
  Frauenfelder, Fujimoto, and Krauthammer}]{m2tr}
Farhad Nooralahzadeh, Nicolas Perez~Gonzalez, Thomas Frauenfelder, Koji
  Fujimoto, and Michael Krauthammer. 2021.
\newblock \href {https://doi.org/10.18653/v1/2021.findings-emnlp.241}
  {Progressive transformer-based generation of radiology reports}.
\newblock In \emph{Findings of the Association for Computational Linguistics:
  EMNLP 2021}, pages 2824--2832, Punta Cana, Dominican Republic. Association
  for Computational Linguistics.

\bibitem[{Papineni et~al.(2002)Papineni, Roukos, Ward, and Zhu}]{bleu}
Kishore Papineni, Salim Roukos, Todd Ward, and Wei-Jing Zhu. 2002.
\newblock \href {https://doi.org/10.3115/1073083.1073135} {{B}leu: a method for
  automatic evaluation of machine translation}.
\newblock In \emph{Proceedings of the 40th Annual Meeting of the Association
  for Computational Linguistics}, pages 311--318, Philadelphia, Pennsylvania,
  USA. Association for Computational Linguistics.

\bibitem[{Qin and Song(2022)}]{cmm-rl}
Han Qin and Yan Song. 2022.
\newblock \href {https://aclanthology.org/2022.findings-acl.38} {Reinforced
  cross-modal alignment for radiology report generation}.
\newblock In \emph{Findings of the Association for Computational Linguistics:
  {ACL} 2022, Dublin, Ireland, May 22-27, 2022}, pages 448--458. Association
  for Computational Linguistics.

\bibitem[{Rennie et~al.(2017)Rennie, Marcheret, Mroueh, Ross, and
  Goel}]{att2in}
Steven~J. Rennie, Etienne Marcheret, Youssef Mroueh, Jerret Ross, and Vaibhava
  Goel. 2017.
\newblock \href {https://doi.org/10.1109/CVPR.2017.131} {Self-critical sequence
  training for image captioning}.
\newblock In \emph{2017 {IEEE} Conference on Computer Vision and Pattern
  Recognition, {CVPR} 2017, Honolulu, HI, USA, July 21-26, 2017}, pages
  1179--1195. {IEEE} Computer Society.

\bibitem[{Smit et~al.(2020)Smit, Jain, Rajpurkar, Pareek, Ng, and
  Lungren}]{chexbert}
Akshay Smit, Saahil Jain, Pranav Rajpurkar, Anuj Pareek, Andrew Ng, and Matthew
  Lungren. 2020.
\newblock \href {https://doi.org/10.18653/v1/2020.emnlp-main.117} {Combining
  automatic labelers and expert annotations for accurate radiology report
  labeling using {BERT}}.
\newblock In \emph{Proceedings of the 2020 Conference on Empirical Methods in
  Natural Language Processing (EMNLP)}, pages 1500--1519, Online. Association
  for Computational Linguistics.

\bibitem[{Song et~al.(2022)Song, Zhang, Ji, Liu, and Wei}]{cmca}
Xiao Song, Xiaodan Zhang, Junzhong Ji, Ying Liu, and Pengxu Wei. 2022.
\newblock \href {https://aclanthology.org/2022.coling-1.210} {Cross-modal
  contrastive attention model for medical report generation}.
\newblock In \emph{Proceedings of the 29th International Conference on
  Computational Linguistics}, pages 2388--2397, Gyeongju, Republic of Korea.
  International Committee on Computational Linguistics.

\bibitem[{Su et~al.(2021{\natexlab{a}})Su, Vandyke, Wang, Fang, and
  Collier}]{plan_then_generate}
Yixuan Su, David Vandyke, Sihui Wang, Yimai Fang, and Nigel Collier.
  2021{\natexlab{a}}.
\newblock \href {https://doi.org/10.18653/v1/2021.findings-emnlp.76}
  {Plan-then-generate: Controlled data-to-text generation via planning}.
\newblock In \emph{Findings of the Association for Computational Linguistics:
  EMNLP 2021}, pages 895--909, Punta Cana, Dominican Republic. Association for
  Computational Linguistics.

\bibitem[{Su et~al.(2021{\natexlab{b}})Su, Wang, Cai, Baker, Korhonen, and
  Collier}]{style_pmi}
Yixuan Su, Yan Wang, Deng Cai, Simon Baker, Anna Korhonen, and Nigel Collier.
  2021{\natexlab{b}}.
\newblock \href {https://doi.org/10.1109/TASLP.2021.3087948}
  {Prototype-to-style: Dialogue generation with style-aware editing on
  retrieval memory}.
\newblock \emph{IEEE/ACM Transactions on Audio, Speech, and Language
  Processing}, 29:2152--2161.

\bibitem[{Vaswani et~al.(2017)Vaswani, Shazeer, Parmar, Uszkoreit, Jones,
  Gomez, Kaiser, and Polosukhin}]{Transformer}
Ashish Vaswani, Noam Shazeer, Niki Parmar, Jakob Uszkoreit, Llion Jones,
  Aidan~N. Gomez, \L{}ukasz Kaiser, and Illia Polosukhin. 2017.
\newblock Attention is all you need.
\newblock In \emph{Proceedings of the 31st International Conference on Neural
  Information Processing Systems}, NIPS'17, page 6000–6010, Red Hook, NY,
  USA. Curran Associates Inc.

\bibitem[{Vinyals et~al.(2015)Vinyals, Toshev, Bengio, and Erhan}]{ST}
Oriol Vinyals, Alexander Toshev, Samy Bengio, and Dumitru Erhan. 2015.
\newblock \href
  {http://dblp.uni-trier.de/db/conf/cvpr/cvpr2015.html#VinyalsTBE15} {Show and
  tell: A neural image caption generator.}
\newblock In \emph{CVPR}, pages 3156--3164. IEEE Computer Society.

\bibitem[{Williams(1992)}]{reinforce}
Ronald~J. Williams. 1992.
\newblock \href {https://doi.org/10.1007/BF00992696} {Simple statistical
  gradient-following algorithms for connectionist reinforcement learning}.
\newblock \emph{Mach. Learn.}, 8(3–4):229–256.

\bibitem[{Wolf et~al.(2020)Wolf, Debut, Sanh, Chaumond, Delangue, Moi, Cistac,
  Rault, Louf, Funtowicz, Davison, Shleifer, von Platen, Ma, Jernite, Plu, Xu,
  Le~Scao, Gugger, Drame, Lhoest, and Rush}]{huggingface}
Thomas Wolf, Lysandre Debut, Victor Sanh, Julien Chaumond, Clement Delangue,
  Anthony Moi, Pierric Cistac, Tim Rault, Remi Louf, Morgan Funtowicz, Joe
  Davison, Sam Shleifer, Patrick von Platen, Clara Ma, Yacine Jernite, Julien
  Plu, Canwen Xu, Teven Le~Scao, Sylvain Gugger, Mariama Drame, Quentin Lhoest,
  and Alexander Rush. 2020.
\newblock \href {https://doi.org/10.18653/v1/2020.emnlp-demos.6} {Transformers:
  State-of-the-art natural language processing}.
\newblock In \emph{Proceedings of the 2020 Conference on Empirical Methods in
  Natural Language Processing: System Demonstrations}, pages 38--45, Online.
  Association for Computational Linguistics.

\bibitem[{Yang et~al.(2021)Yang, Wu, Ge, Zhou, and Xiao}]{mia}
Shuxin Yang, Xian Wu, Shen Ge, Shaohua~Kevin Zhou, and Li~Xiao. 2021.
\newblock \href {http://arxiv.org/abs/2112.15009} {Knowledge matters: Radiology
  report generation with general and specific knowledge}.
\newblock \emph{CoRR}, abs/2112.15009.

\bibitem[{You et~al.(2021)You, Liu, Ge, Xie, Zhang, and Wu}]{aligntransformer}
Di~You, Fenglin Liu, Shen Ge, Xiaoxia Xie, Jing Zhang, and Xian Wu. 2021.
\newblock \href {https://doi.org/10.1007/978-3-030-87199-4\_7}
  {Aligntransformer: Hierarchical alignment of visual regions and disease tags
  for medical report generation}.
\newblock In \emph{Medical Image Computing and Computer Assisted Intervention -
  {MICCAI} 2021 - 24th International Conference, Strasbourg, France, September
  27 - October 1, 2021, Proceedings, Part {III}}, volume 12903 of \emph{Lecture
  Notes in Computer Science}, pages 72--82. Springer.

\end{thebibliography}
\bibliographystyle{acl_natbib}
\appendix
\section{Appendices}
\subsection{Observation Statistics}\label{appendix:observation}
There are 14 categories of observations: \textit{No Finding}, \textit{Enlarged Cardiomediastinum}, \textit{Cardiomegaly}, \textit{Lung Lesion}, \textit{Lung Opacity}, \textit{Edema}, \textit{Consolidation}, \textit{Pneumonia}, \textit{Atelectasis}, \textit{Pneumothorax}, \textit{Pleural Effusion}, \textit{Pleural Other}, \textit{Fracture}, and \textit{Support Devices}. Table \ref{table: obs_stat} lists the observation distributions annotated by CheXbert\cite{chexbert} in the train/valid/test split of two benchmarks.
\begin{table}[hpbt]
    \centering
    \resizebox{\linewidth}{!}{
    \begin{tabular}{l|l|l}
    \Xhline{2\arrayrulewidth}
    \textbf{\#Observation} & \textbf{\textsc{IU X-ray}} & \textbf{\textsc{MIMIC-CXR}} \\\hline
    \textit{No Finding}/POS & 744/108/318 & 64,677/514/229 \\
    \textit{No Finding}/NEG & 1,325/188/272 & 206,133/1,616/3,629 \\
    \hline
    \textit{Cardiomegaly}/POS & 244/38/61 & 70,561/514/1,602 \\
    \textit{Cardiomegaly}/NEG & 1,375/198/386 & 85,448/714/801 \\
    \hline
    \textit{Pleural Effusion}/POS & 60/13/15 & 56,972/477/1,379 \\
    \textit{Pleural Effusion}/NEG & 1,559/230/452 & 170,989/1,310/1,763 \\
    \hline
    \textit{Pneumothorax}/POS & 9/2/5 & 8,707/62/106 \\
    \textit{Pneumothorax}/NEG & 1,528/231/449 & 190,356/1,495/2,338 \\
    \hline
    \textit{Enlarged Card.}/POS & 159/29/28 & 49,806/413/1,140 \\
    \textit{Enlarged Card.}/NEG & 1,200/161/384 & 129,360/1,006/868 \\
    \hline
    \textit{Consolidation}/POS & 17/1/3 & 14,449/119/384 \\
    \textit{Consolidation}/NEG & 763/117/210 & 97,197/788/964 \\
    \hline
    \textit{Lung Opacity}/POS & 295/35/57 & 67,714/497/1,448 \\
    \textit{Lung Opacity}/NEG & 331/49/82 & 8,157/73/125 \\
    \hline
    \textit{Fracture}/POS & 84/6/15 & 11,070/59/232 \\
    \textit{Fracture}/NEG & 137/22/50 & 9,632/72/53 \\
    \hline
    \textit{Lung Lesion}/POS & 85/14/17 & 11,717/123/300 \\
    \textit{Lung Lesion}/NEG & 92/10/30 & 1,972/21/11 \\
    \hline
    \textit{Edema}/POS & 28/2/7 & 33,034/257/899 \\
    \textit{Edema}/NEG & 119/17/31 & 51,639/409/669 \\
    \hline
    \textit{Atelectasis}/POS & 143/15/37 & 68,273/515/1,210 \\
    \textit{Atelectasis}/NEG & 3/0/0 & 563/5/9 \\
    \hline
    \textit{Support Devices}/POS & 89/20/16 & 60,455/450/1,358 \\
    \textit{Support Devices}/NEG & 1/0/0 & 1,081/7/11 \\
    \hline
    \textit{Pneumonia}/POS & 20/2/1 & 23,945/184/503 \\
    \textit{Pneumonia}/NEG & 68/9/25 & 21,976/165/411 \\
    \hline
    \textit{Pleural Other}/POS & 32/4/7 & 7,296/70/184\\
    \textit{Pleural Other}/NEG & 0/0/0 & 63/0/0\\
    \Xhline{2\arrayrulewidth}
    \end{tabular}}
    \caption{Observation distribution in train/valid/test split of two benchmarks. \textit{Enlarged Card.} refers to \textit{Enlarged Cardiomediastinum}.}
    \label{table: obs_stat}
\end{table}
\subsection{Observation-aware N-grams}\label{appendix:ngram}
\begin{figure}[hpbt]
\centering
\includegraphics[width=1.0\linewidth]{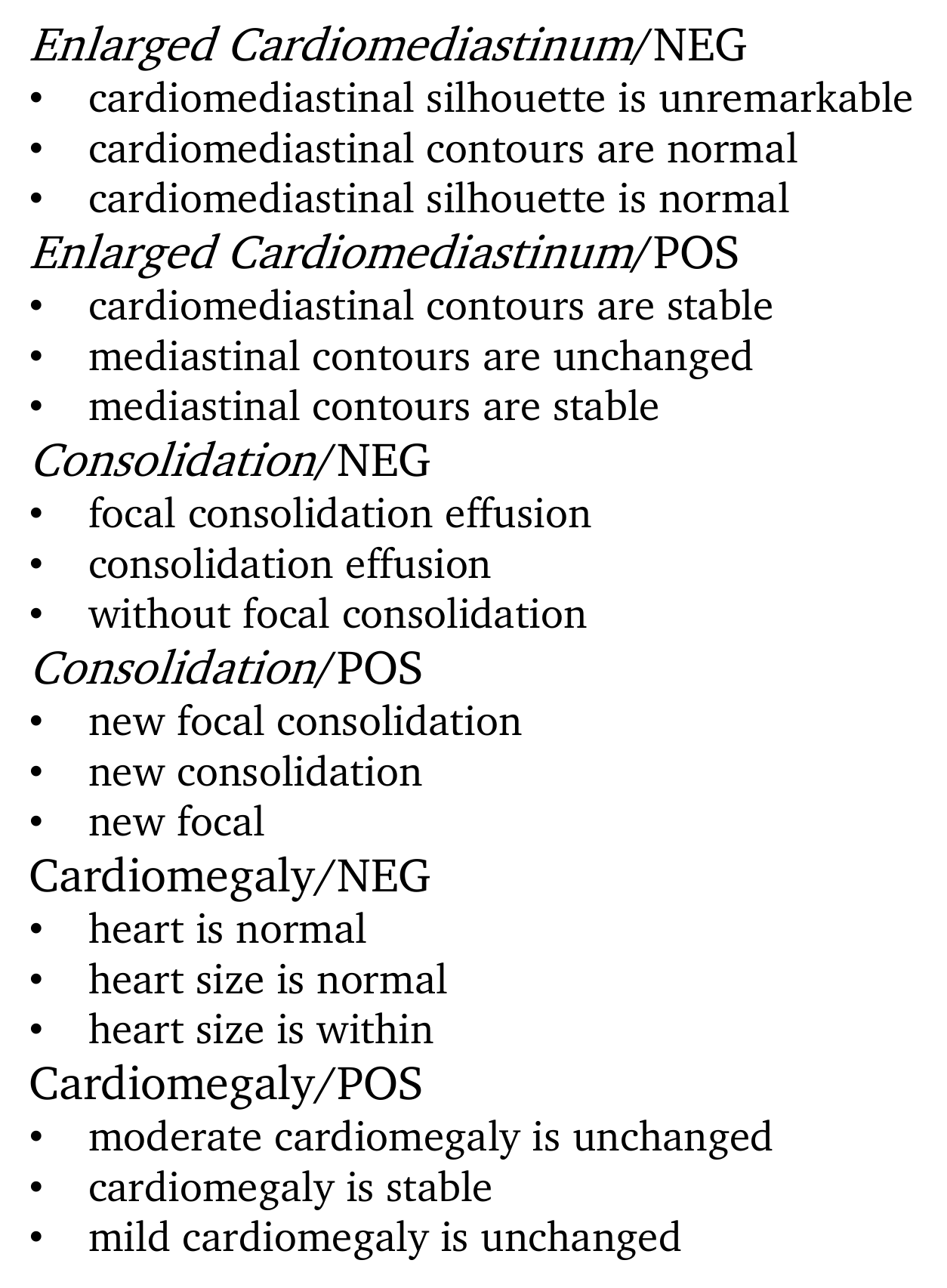}
\caption{Observation-aware n-grams.}
\label{figure: obs_ngram_example}
\end{figure}

Here are some of the observation-aware n-grams we use in our experiments, as shown in Figure \ref{figure: obs_ngram_example}. These categories are \textit{Enlarged Cardiomediastinum}, \textit{Consolidation}, and \textit{Cardiomegaly}.
\end{document}